\PassOptionsToPackage{table}{xcolor}
\documentclass{fairmeta}

\usepackage{amsmath,amsfonts,bm}

\def\eqref#1{equation~\ref{#1}}

\def\1{\bm{1}}

\def\vh{{\bm{h}}}

\def\vx{{\bm{x}}}

\def\vz{{\bm{z}}}

\DeclareMathAlphabet{\mathsfit}{\encodingdefault}{\sfdefault}{m}{sl}
\SetMathAlphabet{\mathsfit}{bold}{\encodingdefault}{\sfdefault}{bx}{n}

\usepackage{amssymb}
\usepackage{float}
\usepackage{wrapfig}
\usepackage{siunitx}
\usepackage{flafter}
\usepackage{needspace}
\usepackage{xspace}
\usepackage{xurl}

\newcommand{\method}{DEdit\xspace}
\newcommand{\proposalmix}{\textsc{ProposalMix}\xspace}
\newcommand{\pass}[1]{P#1}
\newcommand{\masktok}{\text{\textsc{mask}}}
\hypersetup{
  pdftitle={DEdit: Iterative Draft Editing for Speculative Decoding},
  pdfauthor={Longxuan Yu, Bingsen Chen, Peng Shi, Dongkyu Lee, Yi Xiang, Hideo Kobayashi, Sheng Zhang, Shuaichen Chang, Xing Niu, Zhuoyan Xu, Greg Ver Steeg, Jiarong Jiang}
}

\title{DEdit: Iterative Draft Editing for Speculative Decoding}

\author[1,2,*,\dagger]{Longxuan Yu}
\author[2,3,*,\dagger]{Bingsen Chen}
\author[2]{Peng Shi}
\author[2]{Dongkyu Lee}
\author[2]{Yi Xiang}
\author[2]{Hideo Kobayashi}
\author[2]{Sheng Zhang}
\author[2]{Shuaichen Chang}
\author[2]{Xing Niu}
\author[2]{Zhuoyan Xu}
\author[1]{Greg Ver Steeg}
\author[2]{Jiarong Jiang}

\affiliation[1]{University of California, Riverside}
\affiliation[2]{Amazon Web Services}
\affiliation[3]{New York University}
\contribution[*]{Work done during an internship at Amazon Web Services}
\contribution[\dagger]{Equal contribution}

\authoremails{\email{ylong030@ucr.edu}, \email{bale.chen@nyu.edu}, \email{penshi@amazon.com}}

\abstract{
Speculative decoding accelerates autoregressive LLMs by having a lightweight drafter propose tokens that the target model verifies in parallel. Diffusion-based drafters further reduce drafting latency by proposing multiple tokens at once. However, these tokens are predicted independently, so a single early error causes prefix verification to discard the rest of the draft, even when it contains useful downstream predictions. We introduce \method, a diffusion-based drafter that can not only draft by conventional parallel unmasking but also iteratively edit its draft through token-to-token predictions. Through editing, later predictions can serve as bidirectional context for repairing earlier errors and extending the accepted prefix.
To teach the model to repair errors while preserving correct predictions, we propose \proposalmix, a training scheme that mixes draft predictions with ground-truth tokens based on first-pass confidence during training.  Across seven benchmarks on Qwen3-4B and Qwen3-8B, \method achieves the highest macro-average token acceptance and speedup among the evaluated drafters, reaching macro-average speedups of $5.72\times$ and $5.97\times$ over autoregressive generation under greedy decoding, respectively. Further analysis shows that acceptance improves with more editing passes and wider drafting windows, and that \proposalmix halves harmful edits that shorten the accepted prefix. Moreover, restricting the editor to causal attention lowers acceptance, especially on highly predictable outputs, indicating that future context is a key source of these gains.
}

\begin{document}
\raggedbottom
\maketitle

\section{Introduction}
\label{sec:introduction}

\setlength{\intextsep}{4pt}  
\setlength{\columnsep}{12pt} 

\begin{wrapfigure}{r}{0.48\textwidth}
  \centering
  \vspace{-6pt} 
  \includegraphics[width=0.9\linewidth]
    {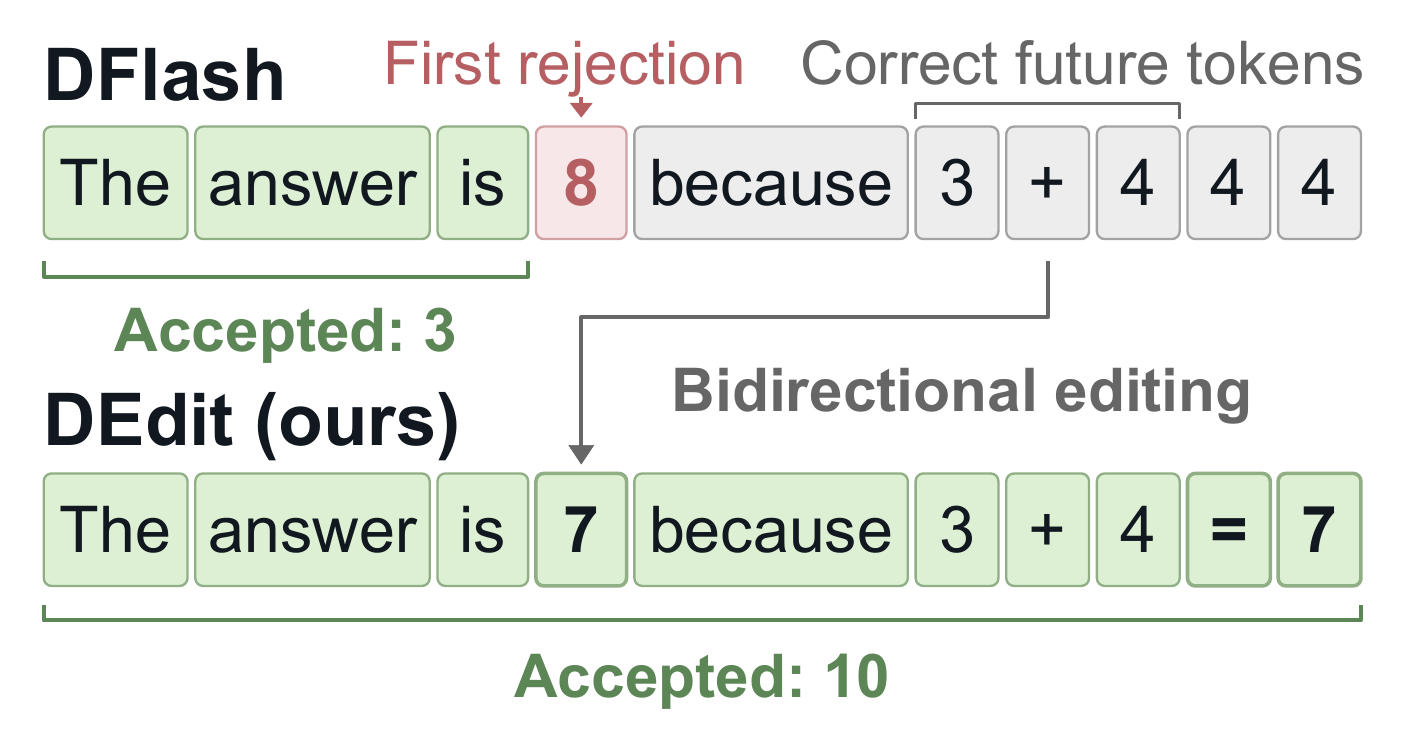}
  \caption{Bidirectional editing before verification extends the accepted prefix.}
  \label{fig:teaser}
\end{wrapfigure}

Autoregressive large language models (LLMs) generate text one token at a time, creating a sequential bottleneck for inference speed. Speculative decoding mitigates this bottleneck by using a lightweight drafter to propose multiple tokens that the target model verifies in a single forward pass~\citep{leviathan2023fast,chen2023speculative}. Its efficiency depends on both the number of accepted tokens and the cost of producing each proposal. However, autoregressive drafters themselves generate proposals sequentially, retaining the overhead of repeated forward passes. To reduce this overhead, parallel drafting, particularly diffusion-based drafting, has been widely adopted in recent work on speculative decoding~\citep{chen2026dflash,huang2026domino,cheng2026dspark,wang2026xpress}.

This shift to parallel drafting introduces a different challenge. Because multiple draft positions are predicted without conditioning on one another, proposal quality tends to degrade at later positions, causing acceptance rates to decay. Existing methods address this limitation by reintroducing causal dependencies among draft tokens to improve proposal quality~\citep{huang2026domino,cheng2026dspark,wang2026xpress}. We take a different view. The lack of causal dependence also means that an error at an early position does not necessarily invalidate later predictions. In fact, later candidates can still be correct even beyond the first rejected token. However, standard prefix verification discards this informative suffix entirely. We therefore ask whether these otherwise wasted parallel predictions can instead be used to revise earlier errors and extend the accepted prefix before target verification.

We introduce \textbf{\method}, a diffusion-based drafter that generates an initial proposal from masked inputs and then iteratively refines all positions in parallel using bidirectional context. This allows later predictions to help repair earlier errors before verification, after which the target model verifies only the final proposal using the standard procedure. Figure~\ref{fig:teaser} illustrates how editing can correct both an early mismatch and the downstream continuation, extending the accepted prefix.

Training \method requires inputs that resemble imperfect first-pass drafts while retaining useful predictions that should not be overwritten. We thus design a novel training algorithm,  \proposalmix, which constructs training proposals by mixing first-pass draft predictions with ground-truth tokens according to confidence. This creates partially correct proposals with both reliable context and realistic errors, teaching the editor when to preserve a prediction and when to revise it.

Our analyses show that the drafter improves token acceptance through inference-time edits beyond the single supervised editing step. Editing also makes wider proposal windows more useful: extending the window from 16 to 32 tokens raises acceptance by 10.3\% after editing, compared with 2.1\% for DFlash~\citep{chen2026dflash}. \proposalmix teaches the editor to avoid harmful edits, halving the rounds in which editing shortens the accepted prefix. Controlled comparisons indicate that editing gains rely on future proposal context, which causal correctors such as Domino~\citep{huang2026domino}, restricted to prefix context, cannot use: hiding later tokens from the editor reduces acceptance on every task we evaluate. This benefit is largest on predictable outputs: on reasoning, summary, and synthetic copy-heavy tasks, \method's acceptance advantage over Domino exceeds that on the standard seven-benchmark suite and is largest on the copy-heavy task.

Across seven benchmarks under greedy and stochastic decoding, \method achieves the highest macro-average token acceptance among the evaluated drafters on both Qwen3-4B and Qwen3-8B. Editing trades additional drafter computation for longer accepted prefixes, so its speed benefit depends on how much drafting cost the execution backend exposes. With CUDA Graph drafting and eager target verification, where the CPU dispatches target operations while the drafter runs, \method also achieves the highest macro-average speedup, reaching $5.72\times$ on Qwen3-4B and $5.97\times$ on Qwen3-8B under greedy decoding.

Our contributions are:
\begin{itemize}
    \item We introduce \method, a bidirectional drafter that generates and iteratively edits complete proposals before verification. Its acceptance improves with additional inference-time edits and continues to increase as proposal windows expand from 16 to 32 tokens.

    \item We develop joint proposal-and-edit training with \proposalmix, which teaches the drafter to preserve correct tokens and revise incorrect ones in a way that aligns editing with the prefix-acceptance objective of speculative decoding.

    \item We provide controlled evidence that access to future proposal context is a key source of the gains from bidirectional editing, revealing a capability unavailable to causal correction methods restricted to prefix context.
\end{itemize}

\section{Preliminaries}
\label{sec:preliminaries}

\subsection{Speculative Decoding}

Speculative decoding accelerates generation from a frozen autoregressive
target model $p$ using an efficient
drafter $q$~\citep{leviathan2023fast,chen2023speculative}. Each decoding
round uses a proposal window of width $W$, comprising one committed anchor
token $y_0$ and $W-1$ unverified candidate tokens. The anchor is the last
token of the current output prefix and remains fixed during drafting and
editing. The target evaluates the window in a single forward pass and
accepts the longest candidate prefix consistent with its own predictions;
the first mismatch terminates acceptance, and the remaining candidates are
discarded regardless of matches at later positions. If $a$ candidates are
accepted, verification appends them and one additional target token, which
becomes the anchor for the next round, advancing generation by $a+1$
tokens. We denote the expected token acceptance per round by
$\tau=\mathbb{E}[a+1]$. Because \method changes only proposal generation
and retains this verification protocol, its improvements appear directly
as a higher $\tau$.

\subsection{Parallel Diffusion Drafting}

Parallel diffusion drafters predict candidate blocks from masked inputs.
We consider the target-conditioned, single-pass formulation exemplified by
DFlash~\citep{chen2026dflash}.
Given decoded context $\vx$ and target
features $\vh(\vx)$, the drafter predicts all masked positions in
$\vz^{(0)}=[y_0,\masktok,\ldots,\masktok]$ simultaneously:
\[
q_{\theta,i}^{(1)}(v)
  := q_{\theta}\!\left(
    Y_i=v \mid \vx,\vh(\vx),\vz^{(0)}\right),
\qquad
\hat{y}_i^{(1)}
  = \arg\max_{v\in\mathcal{V}}q_{\theta,i}^{(1)}(v),
\quad i=1,\ldots,W-1.
\]
The predictions form the initial proposal
$\vz^{(1)}=[y_0,\hat{y}_1^{(1)},\ldots,\hat{y}_{W-1}^{(1)}]$.
\method uses this proposal as the starting point for bidirectional
editing before target verification.

\section{\method}
\label{sec:method}

\subsection{Overview}
\label{sec:method_overview}

\begin{figure}[!t]
  \centering
  \includegraphics[width=0.8\linewidth]
    {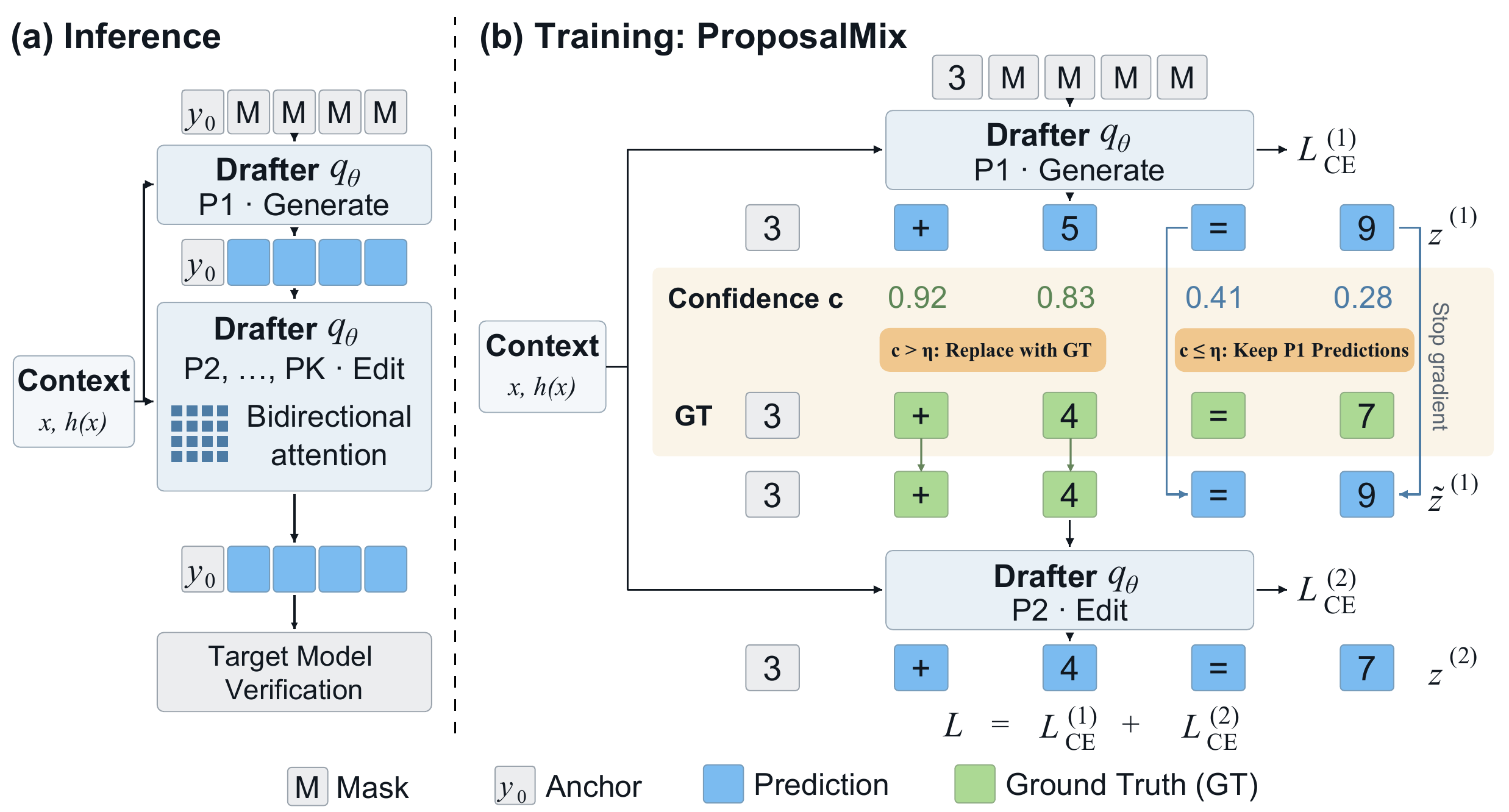}
  \caption{\textbf{\method inference and training.}
  \textbf{(a)} At inference time, drafter generates and edits its proposal tokens through $K$ passes, sending only the final proposal to the target model for verification.
  \textbf{(b)} \proposalmix constructs \pass{2} inputs for joint proposal-and-edit training.}
  \label{fig:method_overview}
\end{figure}

\method uses a diffusion-based drafter to not only generate, but also \textit{iteratively
edit proposals} before target verification.
We first describe the iterative editing procedure, explaining how each
pass uses the preceding proposal as bidirectional context to update
candidate tokens (Section~\ref{sec:iterative_editing}).
We then introduce joint proposal-and-edit training and \proposalmix,
which constructs editing inputs that teach the drafter to revise imperfect proposals while preserving
correct predictions (Section~\ref{sec:proposal_mix}).
Finally, we describe how the trained editor is adapted to larger proposal
windows through continued training
(Section~\ref{sec:window_curriculum}).
Figure~\ref{fig:method_overview} summarizes the inference and training
pipelines.

\subsection{Iterative Proposal Editing}
\label{sec:iterative_editing}

As illustrated in Figure~\ref{fig:method_overview}(a), \method
applies $K-1$ editing passes to the initial proposal $\vz^{(1)}$
using the same drafter, where $K$ denotes the total number of
drafting passes, including initial proposal generation.
Throughout editing, the decoded context $\vx$, target features
$\vh(\vx)$, and anchored token $y_0$ remain fixed.
At each pass $k\in\{2,\ldots,K\}$, the drafter takes the complete
preceding proposal $\vz^{(k-1)}$ as input and predicts:
\[
q_{\theta,i}^{(k)}(v)
  := q_{\theta}\!\left(
      Y_i=v \mid \vx,\vh(\vx),\vz^{(k-1)}\right),
\qquad
\hat{y}_i^{(k)}
  = \arg\max_{v\in\mathcal{V}}q_{\theta,i}^{(k)}(v),
\quad i=1,\ldots,W-1.
\]
These predictions form the updated proposal
$\vz^{(k)}=[y_0,\hat{y}_1^{(k)},\ldots,\hat{y}_{W-1}^{(k)}]$.

All candidate positions are updated simultaneously from the same
preceding proposal. With bidirectional attention, position $i$ can
condition on both earlier and later candidate tokens, including those
at positions $j>i$. Later predictions can therefore provide context
for correcting earlier errors and extending the accepted prefix.
Every candidate position is eligible for revision at each pass,
without confidence gating.

After $K$ passes, only the final proposal $\vz^{(K)}$ is submitted
to the target verifier. All passes share the same parameters, so
each additional edit requires one drafter forward pass but introduces
neither additional model parameters nor an intermediate target query.

\subsection{Joint Training with \proposalmix}
\label{sec:proposal_mix}

We train a single network for both a proposal pass (\pass{1}) and an
editing pass (\pass{2}), as illustrated in
Figure~\ref{fig:method_overview}(b). The proposal objective trains
generation from masked inputs, while the editing objective teaches
the network to preserve correct tokens and revise errors.
To support editing supervision, \proposalmix supplies ground-truth
tokens at positions selected using first-pass confidence and retains
draft predictions elsewhere.
As the drafter's predictions change during training, confidence-based
selection adapts the editing inputs to its evolving error patterns
(Appendix~\ref{app:proposal_mix_statistics}).

For each position $i\in\{1,\ldots,W-1\}$, we define:
\[
c_i = \max_{v\in\mathcal{V}} q_{\theta,i}^{(1)}(v),
\qquad
m_i = \mathbb{I}[c_i > \eta].
\]
Given the ground-truth target token $y_i^\star$, the mixed input is
\[
\tilde{z}^{(1)}_i =
\begin{cases}
y_i^\star, & m_i=1,\\
\hat{y}_i^{(1)}, & m_i=0.
\end{cases}
\]
The anchored token $y_0$ remains fixed, and we set $\eta=0.5$ across all
experiments. At high-confidence positions ($m_i=1$), the ground-truth token
provides correct context for editing. This leaves correct first-pass
predictions unchanged and replaces high-confidence errors. At the remaining
positions ($m_i=0$), the original predictions are retained, exposing the
editor to model-generated errors as well as correct predictions that it
should preserve.

Ground-truth tokens are used as proposal inputs only during training.
The first-pass argmax predictions and confidence masks used to construct
$\tilde{\vz}^{(1)}$ are detached from the computational graph. During
training, the second-pass distribution is
\[
q_{\theta,i}^{(2)}(v)
  := q_{\theta}\!\left(
    Y_i=v \mid \vx,\vh(\vx),\mathrm{detach}(\tilde{\vz}^{(1)})\right),
\qquad i=1,\ldots,W-1.
\]
The editing loss therefore does not backpropagate through the first-pass
predictions or confidence decisions used to construct its input.

Both passes are supervised across all valid future token positions
$\mathcal{T}\subseteq\{1,\ldots,W-1\}$. We use position-decay weights
$w_i=\exp(-(i-1)/\gamma)$, where $\gamma>0$ controls the decay.
These weights emphasize earlier positions because an early mismatch
prevents later candidates from being accepted, while retaining supervision
for every valid position:
\[
\mathcal{L}_{\mathrm{CE}}^{(k)}
  = -\frac{\sum_{i\in\mathcal{T}}
    w_i\log q_{\theta,i}^{(k)}(y_i^\star)}
    {\sum_{i\in\mathcal{T}}w_i},
\qquad k\in\{1,2\}.
\]
We write $\mathcal{L}_{\mathrm{unmask}}=\mathcal{L}_{\mathrm{CE}}^{(1)}$ for
the proposal pass and $\mathcal{L}_{\mathrm{edit}}=\mathcal{L}_{\mathrm{CE}}^{(2)}$
for the editing pass, and train the shared parameters with the sum of these
separately normalized losses:
\[
\mathcal{L}
  = \mathcal{L}_{\mathrm{unmask}}
   +\mathcal{L}_{\mathrm{edit}}.
\]
We refer to minimizing $\mathcal{L}$ over shared parameters $\theta$ as
\emph{joint training}. It contrasts with \emph{two-stage training}, in which
a separate drafter is first trained with $\mathcal{L}_{\mathrm{unmask}}$
and frozen, and an editor is then trained with $\mathcal{L}_{\mathrm{edit}}$
on the drafter's proposals. Joint training avoids a second model at
inference, and it may expose the editing pass to a wider range of errors,
since the errors of the proposal pass change as it improves during training.
\proposalmix changes only the input to \pass{2}; both passes use the same
target labels, valid positions, and loss weights. Editing supervision covers
both positions supplied with ground-truth tokens and those retaining draft
predictions. At inference time, every editing pass consumes the preceding
model-generated proposal, and passes $k\ge3$ reuse the learned editing
operation without additional training.
Appendix~\ref{app:training_recipe} provides the training hyperparameters.

\subsection{Window Expansion}
\label{sec:window_curriculum}

Bidirectional editing allows earlier positions to use later proposal
tokens as context. Expanding the proposal window therefore provides
both more candidates for verification and additional future context
for editing. However, parallel diffusion drafters can exhibit lower
conditional token acceptance rates at later
positions~\citep{sandler2025specdiff2,inco2026dflash2}. Prior evaluations of
DFlash further show that increasing the draft budget can reduce both acceptance
length and end-to-end speedup~\citep{hu2026jetspec}. To benefit from a
larger window, the drafter must produce useful predictions at the
additional positions and use them when revising earlier ones. A drafter
trained at $W=16$, however, has never been supervised at these positions,
so we continue training it at larger windows to adapt both proposal
generation and editing to the expanded context.

We initialize independent training runs at $W=24$ and $W=32$ from the same
converged checkpoint trained at $W=16$, and continue each for 3,000
optimization steps. We increase the position-decay parameter from
$\gamma=7$ at $W=16$ to $\gamma=11$ at $W=24$ and $\gamma=15$ at $W=32$,
slowing the decay of loss weights at more distant positions.

\section{Experiments}
\label{sec:experiments}

\subsection{Setup}
\label{sec:setup}

\textbf{Models and Benchmarks.}
We evaluate \method on Qwen3-4B and Qwen3-8B~\citep{yang2025qwen3}, the
targets used by DFlash and DSpark, across seven benchmarks: GSM8K~\citep{cobbe2021gsm8k},
MATH-500~\citep{hendrycks2021math,lightman2023let}, and
AIME25~\citep{aime2025} (\textsc{Math});
HumanEval~\citep{chen2021humaneval}, MBPP~\citep{austin2021mbpp}, and
LiveCodeBench~\citep{jain2024livecodebench} (\textsc{Code}); and
MT-Bench~\citep{zheng2023mtbench} (\textsc{Chat}). Main results
(\S\ref{sec:main_results}) use an approximately 800K target-aligned corpus
of Nemotron and CodeAlpaca prompts~\citep{nathawani2025nemotron,chaudhary2023codealpaca};
training-design ablations (\S\ref{sec:ablations}) use a separately prepared
100K corpus. Appendix~\ref{app:training_data} reports their sources and exact statistics.

\textbf{Baselines and Drafter Configurations.}
We compare against DFlash~\citep{chen2026dflash}, a single-pass
block-diffusion drafter; Domino~\citep{huang2026domino}, which adds
lightweight causal correction; and DSpark~\citep{cheng2026dspark}, which
adds a sequential module and confidence-scheduled verification.
At $W=16$, all methods are trained on the same corpus for the same number
of epochs, use a five-layer drafter conditioned on the same target layers,
and share the evaluation protocol. DFlash and \method have identical
architectures; Domino adds a lightweight GRU corrector, and DSpark adds a
low-rank Markov head and a confidence head, which it uses to verify only a
confidence-selected prefix of its draft. Following their original
configurations, Domino and DSpark draft $W$ candidates per window rather
than $W-1$. Main results compare \method at $W=32$ with two editing passes
($K=3$, \pass{3}) against baselines at their native $W=16$.
Section~\ref{sec:ablations} reports window-matched comparisons, and
Appendix~\ref{app:pass_depth} the trade-off across editing passes.

\textbf{Evaluation Protocol.}
We evaluate greedy ($T_p=0$) and stochastic ($T_p=1$) decoding
with a maximum of 2,048 generated tokens, following the verification
protocol in Section~\ref{sec:preliminaries}. Under greedy decoding,
verification accepts candidates matching the target's top-1 predictions;
under stochastic decoding, we follow the DFlash
implementation~\citep{chen2026dflash}, in which the drafter proposes argmax
tokens and verification accepts candidates matching tokens sampled from the
target. Thinking mode is disabled
for both target models, both when generating training responses and
during evaluation.
We report the mean token acceptance per round $\tau$ and the end-to-end
speedup over autoregressive decoding, measured on a single
NVIDIA H100 GPU at batch size 1.
Drafting uses CUDA Graphs and target verification runs in eager mode for
all methods, including the autoregressive baseline;
Appendix~\ref{app:systems} details this execution setting and its effect
on measured latency. Full training configurations are provided
in Appendix~\ref{app:training_recipe}.

\Needspace{12\baselineskip}
\subsection{Main Results}
\label{sec:main_results}

Table~\ref{tab:main_results} reports $\tau$ and end-to-end speedup for both
target models under greedy and stochastic decoding.
We draw three conclusions.

\begin{table}[H]
\centering
\caption{End-to-end speedup and mean token acceptance ($\tau$) under the
evaluation protocol in Section~\ref{sec:setup}; Avg. is the seven-suite macro average.
Baselines operate at $W=16$ (Domino and DSpark draft 16 candidates; DFlash
drafts 15), while \method uses the $W=32$ \pass{3} model
after a 3K-step continuation from $W=16$. Bold marks the best value in
each model/temperature block.}
\label{tab:main_results}
\begingroup
\small
\setlength{\tabcolsep}{2pt}
\resizebox{\linewidth}{!}{%
\begin{tabular}{@{}c l @{\hspace{1.0em}} cc cc cc
  @{\hspace{1.0em}} cc cc cc @{\hspace{1.0em}} cc
  @{\hspace{1.0em}} cc@{}}
\toprule
\multirow{2}{*}{Model} & \multirow{2}{*}{Method}
  & \multicolumn{6}{c@{\hspace{1.0em}}}{\textsc{Math}}
  & \multicolumn{6}{c@{\hspace{1.0em}}}{\textsc{Code}}
  & \multicolumn{2}{c@{\hspace{1.0em}}}{\textsc{Chat}}
  & \multicolumn{2}{c}{\textsc{Overall}} \\
\cmidrule(lr){3-8}\cmidrule(lr){9-14}
\cmidrule(lr){15-16}\cmidrule(lr){17-18}
& & \multicolumn{2}{c}{GSM8K}
  & \multicolumn{2}{c}{MATH-500}
  & \multicolumn{2}{c}{AIME25}
  & \multicolumn{2}{c}{HumanEval}
  & \multicolumn{2}{c}{MBPP}
  & \multicolumn{2}{c}{LCB}
  & \multicolumn{2}{c}{MT-Bench}
  & \multicolumn{2}{c}{Avg.} \\
\midrule
\multicolumn{2}{c}{Temperature: 0}
  & \scriptsize Speedup & \scriptsize $\tau$
  & \scriptsize Speedup & \scriptsize $\tau$
  & \scriptsize Speedup & \scriptsize $\tau$
  & \scriptsize Speedup & \scriptsize $\tau$
  & \scriptsize Speedup & \scriptsize $\tau$
  & \scriptsize Speedup & \scriptsize $\tau$
  & \scriptsize Speedup & \scriptsize $\tau$
  & \scriptsize Speedup & \scriptsize $\tau$ \\
\midrule
\multirow{4}{*}{Qwen3-4B}
  & DFlash
  & 4.99$\times$ & 6.04 & 6.52$\times$ & 7.76
  & 5.82$\times$ & 7.07 & 4.45$\times$ & 5.32
  & 4.08$\times$ & 5.06 & 4.81$\times$ & 5.83
  & 2.98$\times$ & 4.17 & 4.81$\times$ & 5.89 \\
  & DSpark
  & 5.37$\times$ & 7.13 & 6.52$\times$ & 9.02
  & 6.06$\times$ & 8.23 & 4.61$\times$ & 6.23
  & 4.37$\times$ & 6.11 & 5.32$\times$ & 7.14
  & 2.67$\times$ & 4.53 & 4.99$\times$ & 6.91 \\
  & Domino
  & 5.53$\times$ & 6.76 & 7.21$\times$ & 8.87
  & 6.65$\times$ & 8.12 & 4.76$\times$ & 5.84
  & 4.56$\times$ & 5.70 & 5.30$\times$ & 6.39
  & 3.13$\times$ & 4.53 & 5.30$\times$ & 6.60 \\
  & \method \pass{3}
  & \textbf{6.51$\times$} & \textbf{7.75}
  & \textbf{8.18$\times$} & \textbf{10.67}
  & \textbf{6.84$\times$} & \textbf{9.28}
  & \textbf{4.95$\times$} & \textbf{6.48}
  & \textbf{4.71$\times$} & \textbf{6.13}
  & \textbf{5.71$\times$} & \textbf{7.61}
  & \textbf{3.17$\times$} & \textbf{5.11}
  & \textbf{5.72$\times$} & \textbf{7.58} \\
\midrule
\multirow{4}{*}{Qwen3-8B}
  & DFlash
  & 5.05$\times$ & 6.09 & 6.39$\times$ & 7.91
  & 6.20$\times$ & 7.39 & 4.49$\times$ & 5.30
  & 4.48$\times$ & 5.04 & 5.06$\times$ & 6.17
  & 2.81$\times$ & 4.05 & 4.93$\times$ & 5.99 \\
  & DSpark
  & 5.60$\times$ & 7.40 & 6.78$\times$ & 9.42
  & 6.29$\times$ & 8.72 & 4.73$\times$ & 6.44
  & 4.76$\times$ & \textbf{6.23} & 5.38$\times$ & 7.60
  & 2.57$\times$ & 4.57 & 5.16$\times$ & 7.20 \\
  & Domino
  & 5.85$\times$ & 6.86 & 7.66$\times$ & 9.06
  & 7.27$\times$ & 8.72 & 4.85$\times$ & 5.97
  & \textbf{4.90$\times$} & 5.71 & 5.40$\times$ & 6.73
  & 3.00$\times$ & 4.48 & 5.56$\times$ & 6.79 \\
  & \method \pass{3}
  & \textbf{6.46$\times$} & \textbf{7.87}
  & \textbf{8.68$\times$} & \textbf{10.94}
  & \textbf{7.48$\times$} & \textbf{9.98}
  & \textbf{5.08$\times$} & \textbf{6.48}
  & 4.87$\times$ & 6.09
  & \textbf{6.04$\times$} & \textbf{8.00}
  & \textbf{3.16$\times$} & \textbf{4.89}
  & \textbf{5.97$\times$} & \textbf{7.75} \\
\midrule
\multicolumn{2}{c}{Temperature: 1}
  & \scriptsize Speedup & \scriptsize $\tau$
  & \scriptsize Speedup & \scriptsize $\tau$
  & \scriptsize Speedup & \scriptsize $\tau$
  & \scriptsize Speedup & \scriptsize $\tau$
  & \scriptsize Speedup & \scriptsize $\tau$
  & \scriptsize Speedup & \scriptsize $\tau$
  & \scriptsize Speedup & \scriptsize $\tau$
  & \scriptsize Speedup & \scriptsize $\tau$ \\
\midrule
\multirow{4}{*}{Qwen3-4B}
  & DFlash
  & 4.55$\times$ & 5.62 & 5.13$\times$ & 6.38
  & 4.09$\times$ & 4.86 & 4.17$\times$ & 4.82
  & 3.95$\times$ & 4.76 & 4.36$\times$ & 5.35
  & 2.86$\times$ & 3.85 & 4.16$\times$ & 5.09 \\
  & DSpark
  & 4.87$\times$ & 6.52 & 5.31$\times$ & 7.28
  & 3.66$\times$ & 5.19 & 4.27$\times$ & 5.72
  & 4.15$\times$ & \textbf{5.62} & 4.89$\times$ & 6.58
  & 2.50$\times$ & 4.10 & 4.24$\times$ & 5.86 \\
  & Domino
  & 5.28$\times$ & 6.21 & 5.76$\times$ & 7.07
  & 4.28$\times$ & 5.38
  & 4.46$\times$ & 5.40
  & 4.36$\times$ & 5.24 & 4.90$\times$ & 5.89
  & 2.91$\times$ & 4.11 & 4.56$\times$ & 5.61 \\
  & \method \pass{3}
  & \textbf{5.61$\times$} & \textbf{6.96}
  & \textbf{6.37$\times$} & \textbf{8.26}
  & \textbf{4.33$\times$} & \textbf{5.62}
  & \textbf{4.94$\times$} & \textbf{5.86}
  & \textbf{4.67$\times$} & 5.56
  & \textbf{5.22$\times$} & \textbf{6.71}
  & \textbf{3.01$\times$} & \textbf{4.58}
  & \textbf{4.88$\times$} & \textbf{6.22} \\
\midrule
\multirow{4}{*}{Qwen3-8B}
  & DFlash
  & 4.48$\times$ & 5.54 & 5.34$\times$ & 6.27
  & 3.74$\times$ & 4.69 & 3.94$\times$ & 4.59
  & 4.03$\times$ & 4.52 & 4.73$\times$ & 5.75
  & 2.53$\times$ & 3.57 & 4.11$\times$ & 4.99 \\
  & DSpark
  & 4.86$\times$ & 6.61 & 5.08$\times$ & 7.18
  & 3.54$\times$ & 5.13 & 4.05$\times$ & \textbf{5.45}
  & 4.14$\times$ & \textbf{5.37} & 5.03$\times$ & 7.07
  & 2.45$\times$ & 4.02 & 4.16$\times$ & 5.83 \\
  & Domino
  & 5.22$\times$ & 6.18 & 5.72$\times$ & 7.01
  & 4.21$\times$ & 5.25 & 4.16$\times$ & 4.99
  & 4.39$\times$ & 5.01 & 5.10$\times$ & 6.22
  & 2.72$\times$ & 3.93 & 4.50$\times$ & 5.51 \\
  & \method \pass{3}
  & \textbf{5.69$\times$} & \textbf{6.94}
  & \textbf{6.47$\times$} & \textbf{8.02}
  & \textbf{4.50$\times$} & \textbf{5.59}
  & \textbf{4.53$\times$} & 5.32
  & \textbf{4.55$\times$} & 5.31
  & \textbf{5.64$\times$} & \textbf{7.18}
  & \textbf{2.89$\times$} & \textbf{4.27}
  & \textbf{4.90$\times$} & \textbf{6.09} \\
\bottomrule
\end{tabular}%
}
\endgroup
\end{table}

\method achieves the highest macro-average $\tau$ in all four target and
decoding settings. Under greedy decoding, it reaches $\tau=7.58$ on Qwen3-4B
and $7.75$ on Qwen3-8B, $9.7\%$ and $7.6\%$ higher than DSpark, the
baseline with the highest $\tau$; under stochastic decoding, the advantage
narrows to $6.1\%$ and $4.5\%$. This gain comes from editing rather than
from a stronger initial proposal: \method shares DFlash's architecture, and
its \pass{1} acceptance is on par with DFlash's
(Table~\ref{tab:ablation_panels}(a)).

The acceptance gain varies across benchmarks. On MATH-500, \method improves
$\tau$ over the strongest baseline by $12$--$18\%$ across the four
settings, and on GSM8K and MT-Bench by $5$--$13\%$. On HumanEval and MBPP,
\method and DSpark are within $4\%$ of each other, and DSpark leads in four
of the eight model--decoding combinations. The benefit of editing is
therefore task-dependent; Section~\ref{sec:visibility_tasks} examines how
it varies with the future context available in a proposal.

In our execution setting (Section~\ref{sec:setup}), the acceptance
advantage also yields the highest macro-average speedup, but by a smaller
margin. \method reaches $5.72\times$ and $5.97\times$ under greedy decoding
and $4.88\times$ and $4.90\times$ under stochastic decoding, $7.0$--$8.9\%$
faster than Domino, the fastest baseline. On Qwen3-4B under greedy decoding,
\method's $\tau$ is $14.8\%$ higher than Domino's, whereas its speedup is
$7.9\%$ higher, because each editing pass adds a drafter forward. Editing
thus trades per-round drafting cost for longer accepted prefixes, and
whether this trade pays off depends on how much drafting cost the execution
backend exposes (Appendix~\ref{app:systems}).

\subsection{Ablation Studies}
\label{sec:ablations}

\method combines three design choices: iterative editing
(Section~\ref{sec:iterative_editing}), joint training with \proposalmix
(Section~\ref{sec:proposal_mix}), and window expansion
(Section~\ref{sec:window_curriculum}). We ablate them on Qwen3-4B under
greedy decoding and report macro-average $\tau$ over the seven benchmarks in
Table~\ref{tab:ablation_panels}. Table~\ref{tab:ablation_panels}(a) varies
the window width and the number of editing passes to test whether wider
windows help and whether this depends on editing;
Table~\ref{tab:ablation_panels}(b) removes \proposalmix, joint training, or
both to measure how each contributes to editing quality.

\begin{table}[H]
\centering
\caption{Window scaling and training ablations on Qwen3-4B.
Both panels report macro-average $\tau$ across seven benchmarks.
\textbf{(a)} Proposal width and editing passes at $T_p=0$; Domino drafts $W$
candidates and the other methods $W-1$.
\textbf{(b)} Training components in a controlled 100K setting: a frozen
DFlash drafter supplies \pass{1}, and a separately trained editor is
evaluated at \pass{2} with $W=16$; $\Delta\tau$ is relative to Full.}
\label{tab:ablation_panels}
\medskip
\footnotesize
\renewcommand{\arraystretch}{1.12}
\sisetup{mode=text,detect-weight=true,detect-family=true}
\begin{minipage}[t]{0.45\linewidth}
\centering
\textbf{(a)} Window scaling
\par\smallskip
\setlength{\tabcolsep}{3.5pt}
\begin{tabular*}{\linewidth}{@{\extracolsep{\fill}}lc *{3}{S[table-format=1.2]}@{}}
\toprule
Method & Pass & {$W=16$} & {$W=24$} & {$W=32$} \\
\midrule
DFlash & \pass{1} & 5.89 & 6.01 & 6.01 \\
Domino & \pass{1} & 6.60 & 6.91 & 6.93 \\
\midrule
\multirow{3}{*}{\method} & \pass{1} & 5.82 & 5.93 & 5.93 \\
 & \pass{2} & 6.63 & 7.02 & 7.12 \\
 & \pass{3} & 6.87 & 7.39 & \bfseries 7.58 \\
\bottomrule
\end{tabular*}
\end{minipage}\hfill
\begin{minipage}[t]{0.52\linewidth}
\centering
\textbf{(b)} Training ablations
\par\smallskip
\setlength{\tabcolsep}{2.5pt}
\renewcommand{\arraystretch}{1.34}
\begin{tabular*}{\linewidth}{@{\extracolsep{\fill}}lcc S[table-format=1.3] S[table-format=-1.3]@{}}
\toprule
Variant & Joint & ProposalMix & {$\tau$} & {$\Delta\tau$} \\
\midrule
\textbf{Full} & $\checkmark$ & $\checkmark$ & \bfseries 5.499 & {---} \\
\midrule
w/o Joint & -- & $\checkmark$ & 5.414 & -0.085 \\
w/o ProposalMix & $\checkmark$ & -- & 5.355 & -0.144 \\
w/o Both & -- & -- & 5.242 & -0.257 \\
\bottomrule
\end{tabular*}
\end{minipage}
\end{table}

Wider windows raise acceptance mainly through editing. Without editing,
extending $W=16$ to $W=32$ barely changes \method's $\tau$
(5.82$\to$5.93), and DFlash and Domino gain only $2.1\%$ and $5.0\%$. With
two editing passes, the same extension raises $\tau$ by $10.3\%$
(6.87$\to$7.58) and, in a separate timing run, speedup from $5.55\times$
to $5.80\times$. Each editing pass also contributes more at wider windows:
the second editing pass (\pass{3}) adds 0.24 at $W=16$ but 0.46 at $W=32$, although it is
never supervised during training. The additional positions of a wider
window are therefore more valuable as context for editing than as extra
candidates for verification.

\proposalmix and joint training both improve editing, with
\proposalmix contributing more. To isolate the training objective,
Table~\ref{tab:ablation_panels}(b) holds the proposal fixed: a frozen
DFlash drafter supplies \pass{1} for all variants, and a separate editor is
trained for each. With joint training, this editor is trained with both
$\mathcal{L}_{\mathrm{unmask}}$ and $\mathcal{L}_{\mathrm{edit}}$;
without it, the editor is trained with $\mathcal{L}_{\mathrm{edit}}$ only,
as in two-stage training (Section~\ref{sec:proposal_mix}). Without
\proposalmix, the editor receives unmodified DFlash proposals. Removing \proposalmix lowers $\tau$ by
0.144 and removing joint training by 0.085, while removing both lowers it
by 0.257. Joint training
helps even though the editor never generates \pass{1} at inference,
suggesting that learning to draft from masked inputs also improves editing.

\section{Analysis}
\label{sec:mechanism_analysis}

Section~\ref{sec:experiments} shows that \method's acceptance gains come
from editing, and that they depend on the training objective and grow with
window width. These aggregate results do not show how the editor achieves
them, which raises two questions. First, \proposalmix raises $\tau$
(Table~\ref{tab:ablation_panels}(b)), but what behavior does it actually
teach the editor (Section~\ref{sec:proposal_mix_analysis})? Second, wider windows help
mainly through editing (Table~\ref{tab:ablation_panels}(a)), suggesting
that the editor uses later proposal tokens as context; does it, and when
does this help most (Section~\ref{sec:why_editing})?

\textbf{Metric.}
We report $A$, the number of accepted draft tokens per round, excluding the
additional target token. To compare drafters with different window widths,
all methods in each analysis are scored on the same number of leading
positions.

\subsection{What Does \proposalmix Teach the Editor?}
\label{sec:proposal_mix_analysis}

\begin{figure}[!t]
  \centering
  \begin{minipage}[t]{0.49\linewidth}
    \vspace{0pt}
    \centering
    \textbf{(a) Effect of \proposalmix}\par\smallskip
    \includegraphics[width=0.8\linewidth]
      {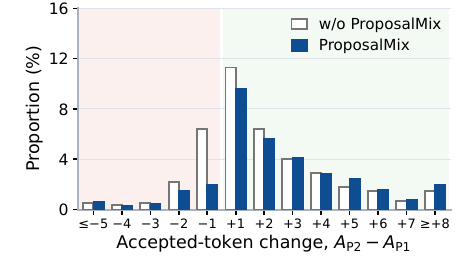}
  \end{minipage}\hfill
  \begin{minipage}[t]{0.49\linewidth}
    \vspace{0pt}
    \centering
    \textbf{(b) Future-context benefit}\par\smallskip
    \includegraphics[width=0.8\linewidth]
      {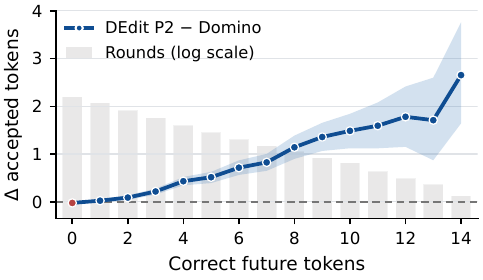}
  \end{minipage}
  \caption{\textbf{Editing outcomes and future context on Qwen3-4B.}
  \textbf{(a)} Edit-gain distribution $A_{\mathrm{P2}}-A_{\mathrm{P1}}$
  for the Full and w/o \proposalmix editors of
  Table~\ref{tab:ablation_panels}(b). Percentages use all rounds;
  zero-gain rounds are omitted (60.0\% without \proposalmix,
  66.5\% with it).
  \textbf{(b)} $W=16$ \method{} \pass{2}-minus-Domino gain versus correct
  future-token count. Points and bands show response-cluster means and
  95\% cluster-bootstrap CIs; gray bars show the number of rounds per bin
  on a log scale, from 56k to 127. Fully accepted rounds (7.7\%) are
  excluded.}
  \label{fig:editing_mechanisms}
\end{figure}

\proposalmix is designed to teach the editor when to preserve a prediction
and when to revise it (Section~\ref{sec:proposal_mix}). To examine what the
editor actually learns, Figure~\ref{fig:editing_mechanisms}(a) compares the
Full and w/o \proposalmix editors of Table~\ref{tab:ablation_panels}(b),
which edit proposals from the same frozen DFlash drafter, and measures the
change in $A$ from \pass{1} to \pass{2}.

The editor trained with \proposalmix preserves more and revises better.
It shortens the accepted prefix in half as many rounds (9.9\% to 4.7\%).
The share of rounds in
which it extends the prefix stays nearly the same (30.1\% to 28.8\%), but
more of these are large repairs: rounds in which editing extends the prefix
by five or more tokens rise from 5.4\% to 6.7\%. Together, these raise the mean edit gain from 0.687 to 0.797.
\proposalmix therefore improves acceptance by avoiding harmful edits rather
than by editing more often. This matters under prefix verification, where a
single harmful edit can invalidate an otherwise accepted prefix.

\subsection{How Does Editing Use Future Context?}
\label{sec:why_editing}
\label{sec:visibility_tasks}

Prefix verification discards correct predictions after the first mismatch,
which we call \emph{correct future tokens} (the trailing C's in
\texttt{[C,C,W,C,C]}, with C correct and W wrong). A bidirectional editor
can attend to them, whereas a causal corrector cannot. We ask whether
editing uses these tokens, and on which tasks this helps most.

\paragraph{Does editing use future context?}
\method's advantage over a causal corrector grows with the number of
correct future tokens. We apply Domino's causal
corrector~\citep{huang2026domino} and \method's \pass{2} editor to
identical $W=16$ proposals from Domino's backbone; the \method editor comes
from the main experiments and has never seen Domino's proposals.
Figure~\ref{fig:editing_mechanisms}(b) shows that the advantage is
negligible at zero correct future tokens and clearly positive from two
onward. After stratifying by accepted-prefix length, each
additional correct future token is associated with a 0.140-token increase
in the advantage (95\% CI $[0.130,0.150]$). Since both correct the same
proposals, this growth points to future context as the source of the
advantage.

Hiding future tokens from the same editor reduces acceptance. Keeping
\method's weights and \pass{1} proposals fixed, we replace bidirectional
attention within the proposal by a lower-triangular mask. On General,
Table~\ref{tab:visibility-intervention} shows that bidirectional attention
adds 0.222 accepted tokens at \pass{2} and 0.290 at \pass{3}. It both
repairs the first rejected token more often and corrupts the accepted
prefix less often (Appendix~\ref{app:future_visibility}), so future context
supports repair and preservation alike. Domino, a trained causal corrector,
provides a reference: at \pass{2}, causal \method falls below Domino on
General, Reasoning, and Summary, whereas bidirectional \method exceeds it on
every slice. Because the editor was trained with
bidirectional attention, part of this gap may reflect the change in
attention pattern at inference.

\begin{table}[!htb]
  \centering
  \small
  \caption{Fixed-weight visibility intervention. Values are $A$ over the
  first 16 candidate positions, averaged over rounds within each slice;
  General averages the seven suite-level means. Causal and bidirectional
  \method{} share a $W=32$ checkpoint and \pass{1} proposal. Domino ($W=16$)
  takes the first token of this proposal and drafts the remaining 15 with its
  own backbone and corrector. $\Delta$: bidirectional minus causal;
  underlined: below Domino.
  Details in Appendix~\ref{app:context_slices}.}
  \label{tab:visibility-intervention}

  \setlength{\tabcolsep}{5pt}
  \renewcommand{\arraystretch}{1.04}

  \begin{tabular}{@{}lrrrrrrr@{}}
    \toprule
    & & \multicolumn{3}{c}{\method{} \pass{2}}
      & \multicolumn{3}{c}{\method{} \pass{3}} \\
    \cmidrule(lr){3-5}
    \cmidrule(l){6-8}
    Task slice & Domino
      & Causal & Bidir. & $\Delta$
      & Causal & Bidir. & $\Delta$ \\
    \midrule
    General (7-suite avg.)
      & 5.168
      & \underline{4.997} & 5.219 & \textbf{+0.222}
      & 5.192 & 5.482 & \textbf{+0.290} \\
    Reasoning
      & 3.260
      & \underline{3.224} & 3.422 & \textbf{+0.199}
      & 3.371 & 3.624 & \textbf{+0.253} \\
    Summary
      & 3.756
      & \underline{3.718} & 3.913 & \textbf{+0.195}
      & 3.881 & 4.141 & \textbf{+0.260} \\
    Copy-heavy
      & 8.560
      & 9.374 & 10.098 & \textbf{+0.724}
      & 9.819 & 10.717 & \textbf{+0.897} \\
    \bottomrule
  \end{tabular}
\end{table}

\paragraph{When does future context help most?}
The benefit of future context is consistent across tasks and largest when
outputs are highly predictable. Besides General, we evaluate Reasoning and
Summary slices of a visible solution and its summary, and a synthetic
Copy-heavy task that copies an input with one specified change
(Appendix~\ref{app:controlled_tasks}). Bidirectional attention improves
acceptance on every slice at both \pass{2} and \pass{3}. On Reasoning and
Summary, it adds about 0.20 tokens at \pass{2} and 0.25 at \pass{3},
comparable to General. On Copy-heavy, it adds 0.724 and 0.897, about three
times as much. When most of the output is determined by the input, proposals likely
contain more correct future tokens for editing to use.

\FloatBarrier
\Needspace{12\baselineskip}
\section{Related Work}
\label{sec:related}

\paragraph{Drafters for speculative decoding.}
Speculative decoding uses a lightweight drafter to propose tokens that the
target model verifies in parallel~\citep{stern2018blockwise,leviathan2023fast,chen2023speculative}.
Drafters include multi-head predictors~\citep{cai2024medusa,ankner2024hydra},
feature-conditioned autoregressive drafters such as
EAGLE~\citep{li2024eagle,li2024eagle2,li2025eagle3}, and diffusion drafters
that denoise all positions in
parallel~\citep{christopher2025specdiff,sandler2025specdiff2,li2025diffuspec,cheng2025deer,chen2026dflash,zhang2026dflare}.
Diffusion drafters are fast because they predict a whole block in one
forward pass, but each position is predicted without conditioning on the
others. Multi-token prediction~\citep{gloeckle2024better,deepseek2024v3} instead
trains the target itself to predict several future tokens, and its
prediction heads can be reused as drafters. These methods improve how the initial proposal is generated; \method instead
edits the realized proposal before verification.

\paragraph{Dependency recovery for parallel drafts.}
Predicting draft positions independently causes acceptance to degrade at
later positions. Recent methods restore dependencies with
lightweight causal correction~\citep{huang2026domino,cheng2026dspark,zheng2026delsspec,wang2026xpress},
extend such correction to draft trees~\citep{li2026dartree}, or select a
coherent path among candidates~\citep{rusanovsky2026lilicorr}. In causal
correction, each position conditions only on earlier tokens, so correct
predictions later in the draft cannot inform earlier ones. \method instead
revises each position using proposal tokens on both sides.

\paragraph{Iterative refinement and training for editing.}
Iterative refinement is well established in non-autoregressive generation,
from Mask-Predict~\citep{ghazvininejad2019mask} to diffusion language models
that revise generated tokens~\citep{bie2026llada21,chen2026dmax}, and
Speculative Correction~\citep{chen2026speculativecorrection} applies
bidirectional refinement to complete drafts. Using refinement for
speculative drafting further requires training, since edits must extend the
accepted prefix rather than merely improve token-level accuracy.
PARD-2~\citep{an2026pard2} and
VAT~\citep{gu2026vat} align drafter training with target acceptance.
\method brings iterative refinement to speculative drafting, and
\proposalmix trains the editor on partially correct proposals to preserve
reliable predictions under prefix verification.

\section{Conclusion}
\label{sec:conclusion}

We propose \method, a diffusion-based drafter that generates proposal tokens
and iteratively edits them before the target model verifies them.
To train this drafter, we design \proposalmix, a training scheme that teaches
the model to repair errors in its proposals while keeping the correct ones. 
Across seven benchmarks under greedy and stochastic decoding, \method achieves
the highest macro-average token acceptance among the evaluated drafters and,
in our execution setting, the highest macro-average speedup. Our analyses show that
editing uses correct future tokens as context, so the additional positions
of a wider window are more valuable for editing than as extra candidates.

\Needspace{8\baselineskip}
\subsection*{AI Use Statement}
In this work, generative AI tools assisted with manuscript drafting
and language editing and with implementing parts of the experimental
code. The training corpora include target-model-generated
responses, as described in Appendix~\ref{app:training_data}.
The authors take full responsibility for the final manuscript,
experimental code, and reported results.

\subsubsection*{Reproducibility Statement}
All model architectures, training hyperparameters, loss formulations, and evaluation protocols are detailed in Section~\ref{sec:experiments} and Appendix~\ref{app:training_details}. The evaluation datasets used across all seven benchmarks are publicly accessible. Code, checkpoints, and evaluation scripts will be made publicly available to facilitate reproducibility.

\clearpage
\bibliographystyle{assets/plainnat}
\bibliography{references}

\clearpage
\beginappendix

\section{Mechanism and Task Analysis}
\label{app:mechanism_analysis}

\subsection{Qualitative Editing Trace}
\label{app:qualitative}

\begin{figure}[H]
  \centering
  \includegraphics[width=0.9\linewidth]
    {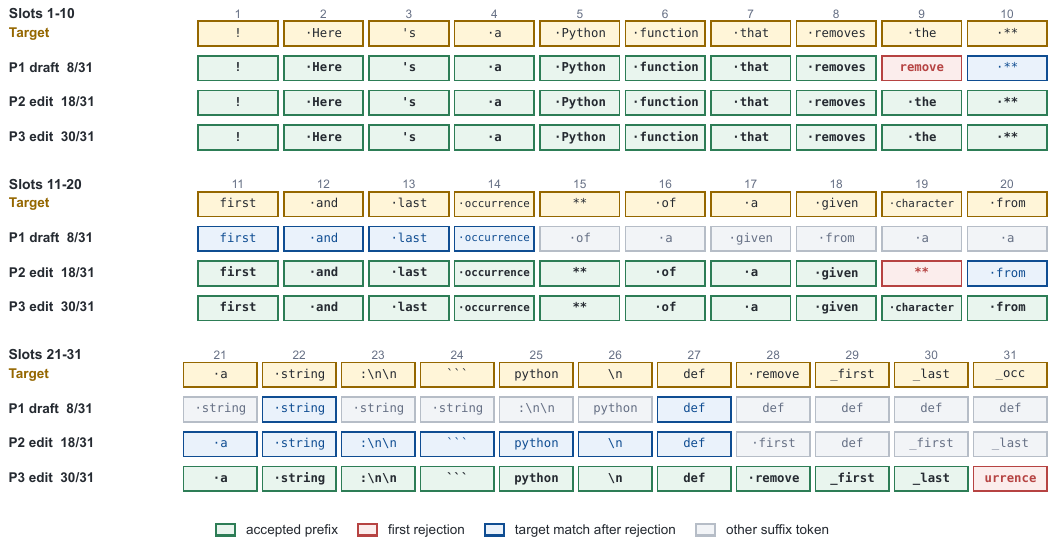}
  \caption{\textbf{A full $W=32$ editing trace.} On an MBPP prompt, successive
  passes extend the accepted prefix from 8 to 18 to 30 of 31 proposal tokens.
  Blue cells already match the target but remain blocked by an earlier error.}
  \label{fig:w32_iterative_trace}
\end{figure}

The trace in Figure~\ref{fig:w32_iterative_trace} uses a fixed-width proposal, so
``insertion'' and ``deletion'' patterns
refer to repairing token shifts under sequence alignment; the tensor length
stays fixed.

Figure~\ref{fig:w32_iterative_trace} shows the complete first verification
round for the $W=32$ \method checkpoint used in the main results. The \pass{1} proposal first
fails at slot 9 even though slots 10--14 already match the target. \pass{2}
repairs the prefix through slot 18 and leaves another correct run at slots
20--27 after its first rejection. \pass{3} repairs the prefix through slot 30
but proposes \texttt{urrence} rather than the target token \texttt{\_occ} at
slot 31. The verifier therefore accepts 30 proposal tokens and emits
\texttt{\_occ} as the next-round anchor. The reported counts are contiguous
accepted-prefix lengths, not token-level accuracies.

\paragraph{Observational audit details.}
The $W=16$ comparison in Section~\ref{sec:why_editing} contains 176,823 decoding
rounds from 2,830 response clusters. \method outperforms Domino on 17.91\% of
rounds, matches on 69.03\%, and loses on 13.06\%. For example,
\texttt{[C,C,W,C,C]} has two correct future tokens after the first rejection.
At a future-token count of 14, the observed gain reaches $+2.654$, although
such high-count rounds are rare.

\FloatBarrier
\subsection{Controlled Task Definitions}
\label{app:controlled_tasks}

We use three reasoning--summary constructions and one copy-heavy task to
control how much useful future information is already visible. These are
mechanism diagnostics, not new task benchmarks. All reasoning prompts are
derived from frozen GSM8K examples, use visible model output with
\texttt{enable\_thinking=false}, and are selected without using drafter
performance. The single-response construction permits separate Reasoning and
Summary measurements; the two-turn and frozen-prefix forms isolate the same
summary-continuation effect under alternative boundaries.

\paragraph{Shared reasoning example.}
The three reasoning--summary formats below use this representative question:
\begin{quote}
\raggedright
\small
Amber, Micah, and Ahito ran 52 miles in total. Amber ran 8 miles. Micah ran
3.5 times what Amber ran. How many miles did Ahito run?
\end{quote}

\paragraph{Two-turn visible reasoning to summary.}
The first turn requests an explicit solution:
\begin{quote}
\raggedright
\small
Write a detailed, checkable solution. Show the essential equations and
intermediate calculations. End with a separate line in exactly this format:
\texttt{Reasoning conclusion: <answer>}.
\end{quote}
The complete target response is then retained in the conversation, and the
second turn asks:
\begin{quote}
\raggedright
\small
Using the solution above, write a concise self-contained solution in 3 to 5
sentences. Keep only the essential calculation steps. End with a separate
line in exactly this format: \texttt{Final answer: <answer>}. Do not mention
the previous response, a reasoning trace, or these instructions.
\end{quote}
Thus the Summary continuation can condition on the exact visible Reasoning
generated in the preceding turn.

\paragraph{Single-response reasoning and summary.}
This construction asks for both stages in one assistant response:
\begin{quote}
\raggedright
\small
Solve the problem and return exactly two plain-text sections in this order.
Do not write any text before the first section label.

\textbf{Reasoning:}\\
Write a detailed, checkable solution with the essential equations and
intermediate calculations.

\textbf{Summary:}\\
Write a concise self-contained solution in 3 to 5 sentences. Keep only the
essential calculation steps and end with a separate line in exactly this
format: \texttt{Final answer: <answer>}.
\end{quote}
A representative frozen target response is:
\begin{quote}
\raggedright
\small
\textbf{Reasoning:}\\
The total distance is 52 miles. Amber ran 8 miles, and Micah ran
$3.5\times 8=28$ miles. Together they ran $8+28=36$ miles, so Ahito ran
$52-36=16$ miles.

\textbf{Summary:}\\
Amber ran 8 miles, and Micah ran $3.5\times 8=28$ miles. Together they ran
36 miles. Ahito therefore ran $52-36=16$ miles.\\
Final answer: 16
\end{quote}
The stage boundary is the literal \texttt{Summary:} delimiter, so proposal
statistics can be accumulated separately before and after it.

\paragraph{Frozen-reasoning summary continuation.}
To isolate Summary generation from variation in the preceding Reasoning, we
freeze the target-produced prefix and begin evaluation after the delimiter:
\begin{quote}
\raggedright
\small
\textbf{Reasoning:}\\
The total distance run by Amber, Micah, and Ahito is 52 miles. Amber ran
8 miles. Micah ran $3.5\times 8=28$ miles. Amber and Micah therefore ran
$8+28=36$ miles, so Ahito ran $52-36=16$ miles.

\textbf{Summary:}
\end{quote}
The expected continuation is:
\begin{quote}
\raggedright
\small
Amber ran 8 miles, and Micah ran $3.5\times 8=28$ miles. Together, Amber and
Micah ran 36 miles. Ahito ran $52-36=16$ miles.\\
Final answer: 16
\end{quote}
Only the Summary suffix is drafted and measured in this construction.

\paragraph{Copy-heavy continuation.}
The CodeXGLUE-derived task exposes the exact old-to-new replacement and asks
for the complete normalized method:
\begin{quote}
\raggedright
\small
Apply the requested bug fix to the normalized Java method.

Replace the exact token sequence
\texttt{<OLD>VAR\_1</OLD>} with
\texttt{<NEW>VAR\_3</NEW>}.

Change nothing else. Return the complete corrected method as a single line
with one space between tokens. Return only the method, with no Markdown, code
fences, labels, or explanation.

{\ttfamily
private boolean METHOD\_1 ( java.lang.String VAR\_1 ) \{ boolean VAR\_2 =
false ; try \{ java.lang.Boolean . METHOD\_2 ( VAR\_3 ) ; VAR\_2 = true ; \}
catch ( TYPE\_1 error ) \{ VAR\_4 . METHOD\_3 ( STRING\_1 ) ; \} return
VAR\_2 ; \}}
\end{quote}
The expected response changes only the supplied token:
\begin{quote}
\raggedright
\small\ttfamily
private boolean METHOD\_1 ( java.lang.String VAR\_3 ) \{ boolean VAR\_2 =
false ; try \{ java.lang.Boolean . METHOD\_2 ( VAR\_3 ) ; VAR\_2 = true ; \}
catch ( TYPE\_1 error ) \{ VAR\_4 . METHOD\_3 ( STRING\_1 ) ; \} return
VAR\_2 ; \}
\end{quote}
Because the gold replacement is present in the prompt, this experiment
measures lexical preservation and local correction, not standard CodeXGLUE
program-repair quality.

\FloatBarrier
\subsection{Task-Slice Evaluation Details}
\label{app:context_slices}

Table~\ref{tab:visibility-intervention} evaluates the visibility intervention on the
seven-suite general workload and the three controlled slices defined above:
a visible reasoning section, its conditioned summary, and a copy-heavy
continuation. All three configurations receive the same target context.
The two \method{} configurations use the same $W=32$ checkpoint, \pass{1}
proposal, and attention implementation, and edit all 31 candidate positions.
Domino drafts 16 candidates at $W=16$ (Section~\ref{sec:setup}); to align its
input with \method, we fix its first candidate to the first token of the
\method \pass{1} proposal, and its backbone and corrector generate the
remaining 15. Domino is a reference rather than part of the fixed-weight
intervention.

For all configurations, we score the accepted prefix, after \pass{2} and
\pass{3} for \method, over the same first 16 candidate positions and exclude
rounds in which fewer than 16 tokens of the target continuation remain. Reasoning and
Summary use a fixed 1,000-prompt GSM8K-derived set; 943 responses satisfy the
required section markers and contribute 26,572 Reasoning and 8,919 Summary
rounds. Copy-heavy uses a fixed 1,000-prompt set and contributes 5,874
rounds. General is the macro average over seven suite-level round means.

\subsection{Fixed-Weight Future-Visibility Counterfactual}
\label{app:future_visibility}

\paragraph{Intervention.}
We directly remove future proposal visibility from the Qwen3-4B $W=32$
\method editor used in the main results while holding its checkpoint, target hidden states, \pass{1}
proposal, verifier, and target-reference trajectory fixed. Both configurations pass an
explicit attention mask over the complete $W=32$ proposal. The bidirectional configuration
exposes all proposal positions, while the causal configuration replaces only
intra-proposal visibility with a lower-triangular mask; all target-prefix
hidden states remain visible. At \pass{2}, both configurations therefore receive exactly the
same \pass{1} input. \pass{3} iteratively edits the proposal produced by its
own \pass{2} configuration. Following the task-slice protocol, reported acceptance
statistics evaluate the first 16 proposal tokens and exclude rounds in which
fewer than 16 tokens of the target continuation remain.

\begin{figure}[H]
  \centering
  \includegraphics[width=0.92\linewidth]
    {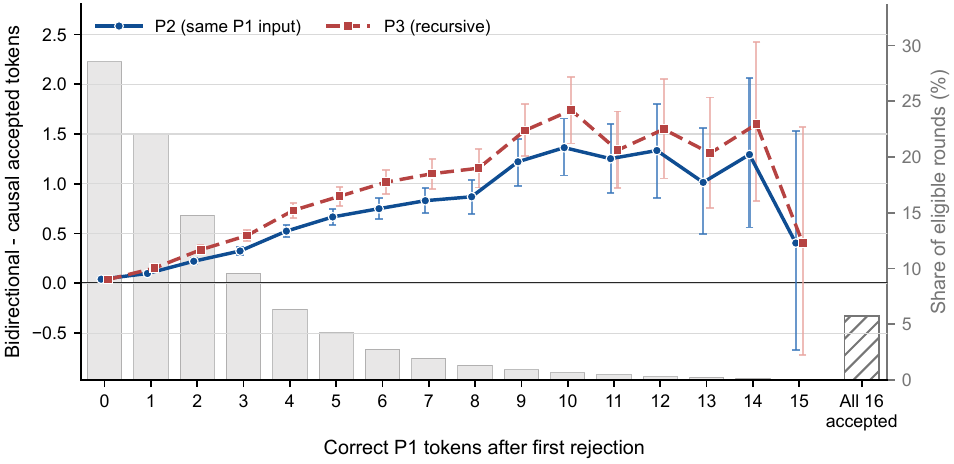}
  \caption{Fixed-weight intervention on future proposal visibility. Lines
  report response-cluster mean bidirectional-minus-causal accepted tokens
  with 95\% response-bootstrap intervals; gray bars report each bin's share
  of 181,205 eligible rounds. \pass{2} compares both masks on the same
  \pass{1} proposal, while \pass{3} applies the next edit to each configuration. The hatched
  bar denotes the 10,382 rounds in which all 16 shared proposal tokens are
  accepted.}
  \label{fig:future_visibility_counterfactual}
\end{figure}

\paragraph{Effect by available future information.}
Averaged over response clusters rather than rounds as in
Table~\ref{tab:visibility-intervention}, bidirectional visibility adds 0.268
accepted tokens
at \pass{2}, with a 95\% interval of $[0.253,0.283]$, and 0.361 at
\pass{3}, with an interval of $[0.343,0.380]$. The effect is positive even
when no correct token follows the first rejection: $+0.041$ at \pass{2} and
$+0.038$ at \pass{3}. It then generally increases with available correct
future tokens, reaching $+0.750$ and $+1.016$ at a count of six.

Task-level results after one and two editing passes are reported in
Table~\ref{tab:visibility-intervention}.

\paragraph{Repair and preservation.}
On General, bidirectional and causal visibility repair the first rejected
token in 27.85\% and 27.11\% of \pass{2} rounds, respectively, while
corrupting the original accepted prefix in 2.82\% and 4.75\%. At \pass{3},
the repair rates are 29.68\% and 27.10\%, and the prefix-corruption rates are
3.14\% and 4.69\%. Thus the editor's bidirectional advantage is visible in
both more frequent repair and better preservation, and its gains are largest
when more correct future tokens follow the first rejection
(Figure~\ref{fig:future_visibility_counterfactual}).

\FloatBarrier
\section{Training Details}
\label{app:training_details}

\subsection{Training Data}
\label{app:training_data}

Table~\ref{tab:training_corpus_composition} reports the source composition
of the target-aligned corpora. Each record pairs a user prompt with a
response generated by the corresponding Qwen3 target with thinking disabled.
Counts refer to released records before training-time tokenization and
sequence-length handling; percentages use each corpus's actual row count.

\paragraph{Main training corpora (800K).}
The main experiments use prompts from the chat, code, and math partitions of
Nemotron Post-Training Dataset v2~\citep{nathawani2025nemotron} and
CodeAlpaca-20k~\citep{chaudhary2023codealpaca}, paired with target-specific
responses. The Qwen3-4B and Qwen3-8B corpora contain 799,864 and
800,000 records, respectively. Their source counts differ by 136 chat
records; code, math, and CodeAlpaca counts are identical.

\paragraph{Controlled training corpus (100K).}
The Qwen3-4B training-design ablations use a 99,987-record corpus.
Its recorded source families are \texttt{nemotron}, \texttt{openr1\_math},
\texttt{opencodeinstruct}, and \texttt{evol\_codealpaca}.
The \texttt{nemotron} records span chat, code, math, and STEM partitions.
Its sources and proportions differ from those of the 800K corpora.

\begin{table}[H]
\centering
\small
\caption{Training-corpus composition: records (percentage of each corpus).
Counts are computed from the released \texttt{source} field and, for 100K,
the joint \texttt{source}/\texttt{split} fields. Nemotron rows denote v2
partitions for 800K and the mirror's \texttt{nemotron} labels for 100K.
A dash denotes no records with that source label.}
\label{tab:training_corpus_composition}
\medskip
\setlength{\tabcolsep}{5pt}
\renewcommand{\arraystretch}{1.08}
\begin{tabular}{@{}lrrr@{}}
\toprule
Source / partition & 800K (Qwen3-4B) & 800K (Qwen3-8B) & 100K (Qwen3-4B) \\
\midrule
Nemotron / chat & 365,375 (45.68\%) & 365,511 (45.69\%) & 28,791 (28.79\%) \\
Nemotron / code & 175,000 (21.88\%) & 175,000 (21.88\%) & 2,195 (2.20\%) \\
Nemotron / math & 239,467 (29.94\%) & 239,467 (29.93\%) & 6,698 (6.70\%) \\
Nemotron / STEM & -- & -- & 23,197 (23.20\%) \\
CodeAlpaca-20k & 20,022 (2.50\%) & 20,022 (2.50\%) & -- \\
OpenR1 Math & -- & -- & 12,429 (12.43\%) \\
OpenCodeInstruct & -- & -- & 19,364 (19.37\%) \\
Evol-CodeAlpaca & -- & -- & 7,313 (7.31\%) \\
\midrule
Total & 799,864 (100\%) & 800,000 (100\%) & 99,987 (100\%) \\
\bottomrule
\end{tabular}
\end{table}

\paragraph{Window-expansion data and settings.}
The $W=24$ and $W=32$ continuations use a fixed 200,000-record sample from
the corresponding target's 800K corpus, selected uniformly without
replacement with seed 42 and retained in source order. This continuation
sample is separate from the 100K training-design corpus. Each continuation
runs 3,000 steps with learning rate $10^{-4}$, block size $W$, decay
$\gamma=11$ and 256 anchors at $W=24$, and $\gamma=15$ and 192 anchors at
$W=32$; other settings follow Table~\ref{tab:training_configs}. DFlash,
Domino, and \method use the same continuation settings.

\subsection{Training Implementation}
\label{app:training_recipe}

Our implementation adopts the DFlash~\citep{chen2026dflash} backbone
and target-feature conditioning scheme. The target model remains frozen
throughout drafter training. Attention over the decoded context is causal,
while attention across candidate positions is bidirectional.
The selected target layers and other training settings are listed in
Table~\ref{tab:training_configs}.

Within each scale, DFlash, Domino~\citep{huang2026domino},
DSpark~\citep{cheng2026dspark}, and \method are trained on the same corpus
for the same number of epochs. Each baseline keeps its original training
objective; DSpark, for example, combines CE and L1 losses with weights 0.1
and 0.9. \method applies position-weighted token CE to every
valid position in both trained passes, in the main and the 100K settings.
Training supervises \pass{1}/\pass{2}; later passes iteratively reuse the
editor at inference without additional training.

\begin{table}[H]
\centering
\small
\setlength{\tabcolsep}{5pt}
\caption{Training configurations for the controlled ablations and main
results. Shared entries span both columns.}
\label{tab:training_configs}
\begin{tabular}{@{}lcc@{}}
\toprule
Setting & Controlled & Main \\
\midrule
Role & Training-design ablations & Main system comparison \\
Target(s) & Qwen3-4B & Qwen3-4B and Qwen3-8B \\
Corpus & 100K target-aligned & 800K Nemotron-based mixture \\
Global batch & 8 & 16 \\
\midrule
Epochs & \multicolumn{2}{c}{6} \\
Optimizer & \multicolumn{2}{c}{AdamW, LR $6\times10^{-4}$, warmup 0.04, clip 1.0} \\
Sequence & \multicolumn{2}{c}{Length 2,048, block 16, 256 anchors} \\
Target layers / decay & \multicolumn{2}{c}{1/9/17/25/33 / 7} \\
\method objective & \multicolumn{2}{c}{Position-weighted \pass{1}+\pass{2} CE} \\
Precision / attention & \multicolumn{2}{c}{BF16 / SDPA} \\
Seed & \multicolumn{2}{c}{42} \\
\bottomrule
\end{tabular}
\end{table}

\subsection{\proposalmix Input Statistics}
\label{app:proposal_mix_statistics}

Table~\ref{tab:training_input_statistics} shows how \proposalmix inputs
change as the proposal pass improves. We replay intermediate checkpoints of
the Full model in Table~\ref{tab:ablation_panels}(b) on a fixed probe set,
using each checkpoint as both \pass{1} and \pass{2}, with $\eta=0.5$. Ground-truth selection includes positions whose first-pass
prediction already matches the target; overwrite counts only positions
whose token changes.

\begin{table}[H]
\centering
\small
\caption{\proposalmix input rates (\%; $\eta=0.5$) when each checkpoint
generates its own \pass{1} on a fixed probe set.
GT selection marks positions chosen for ground-truth conditioning;
overwrite counts those whose tokens change.}
\label{tab:training_input_statistics}
\medskip
\setlength{\tabcolsep}{8pt}
\renewcommand{\arraystretch}{1.08}
\begin{tabular}{@{}lrrr@{}}
\toprule
Step & \pass{1} accuracy & GT selection & Overwrite \\
\midrule
12,221 & 33.19 & 28.10 & 5.79 \\
36,663 & 42.80 & 37.35 & 5.84 \\
73,326 & 49.87 & 44.91 & 5.84 \\
\bottomrule
\end{tabular}
\end{table}

\subsection{Position-Weighted Two-Pass CE}
\label{app:objective}

Let $\mathcal{T}$ contain every valid non-anchor token in the sampled training
blocks. If $r_i\in\{1,\ldots,W-1\}$ is token $i$'s proposal offset, its weight
is
\[
w_i=\exp\!\left(-\frac{r_i-1}{\gamma}\right),
\qquad
Z=\sum_{i\in\mathcal{T}}w_i.
\]
For each trained pass $k\in\{1,2\}$, the implementation computes a separately
normalized cross-entropy,
\[
\mathcal{L}_{\mathrm{CE}}^{(k)}
  =-\frac{1}{Z}\sum_{i\in\mathcal{T}}
    w_i\log q_{\theta,i}^{(k)}(y_i^\star),
\qquad
\mathcal{L}
  =\mathcal{L}_{\mathrm{unmask}}+\mathcal{L}_{\mathrm{edit}},
\]
where $\mathcal{L}_{\mathrm{unmask}}=\mathcal{L}_{\mathrm{CE}}^{(1)}$ and
$\mathcal{L}_{\mathrm{edit}}=\mathcal{L}_{\mathrm{CE}}^{(2)}$.
$W=16$, $W=24$, and $W=32$ use
$\gamma=7,11,15$, respectively. The valid-token mask and positional weights
are the same for both passes; only the \pass{2} input differs. In particular,
\proposalmix does not narrow \pass{2} supervision to uncertain positions.

The positional decay reflects prefix verification: errors near the beginning
of a proposal prevent later tokens from being accepted in the same round.
Supervising the full block still teaches later tokens that can become useful
after an earlier edit. Appendix~\ref{app:pass_depth}
reports how acceptance and speedup vary with the number of editing passes.

\FloatBarrier
\Needspace{8\baselineskip}
\section{Inference and Systems Analysis}
\label{app:inference_systems}

\subsection{Editing-Pass Trade-off}
\label{app:pass_depth}

\begin{figure}[!htb]
  \centering
  \includegraphics[width=0.90\linewidth]{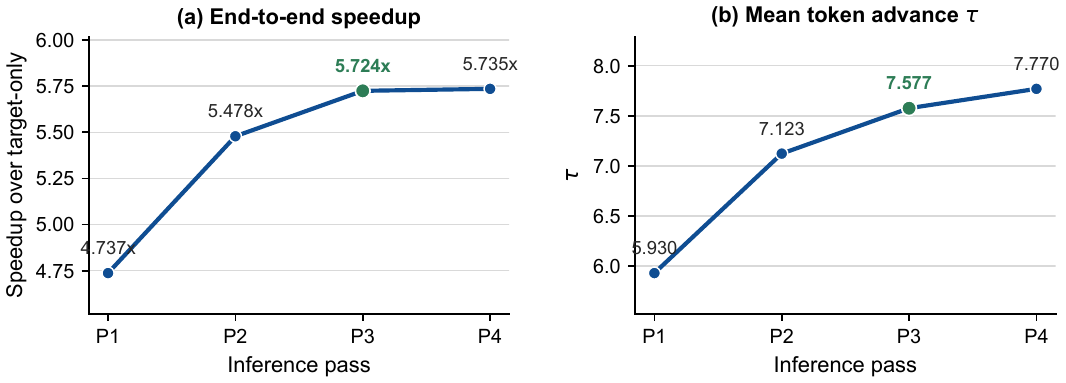}
  \caption{Trade-off across editing passes for the same $W=32$ checkpoint at
  $T_p=0$. Each configuration is evaluated twice with a shared
  autoregressive baseline. Drafting uses CUDA Graphs and target
  verification remains eager. Training supervises \pass{1} and
  \pass{2}; \pass{3} and \pass{4} reuse the trained editor at inference.}
  \label{fig:pass_depth}
\end{figure}

As Figure~\ref{fig:pass_depth} shows, adding a second editing pass (from
\pass{2} to \pass{3}) raises $\tau$ from 7.123
to 7.577 and speedup from $5.478\times$ to $5.724\times$.
\pass{4} further increases $\tau$ to 7.770, while speedup rises
by only 0.19\% to $5.735\times$. Although \pass{4} is the fastest
measured configuration, we use \pass{3} as the default because it
retains 99.8\% of the observed \pass{4} speedup with one fewer
drafter forward per round.

\subsection{Execution Setting}
\label{app:systems}

We evaluate single-request decoding with CUDA Graph-optimized
drafting and eager target verification. CUDA Graph replay reduces
host submission overhead by launching a captured sequence of GPU
operations through one submission. We preallocate fixed-shape
input, position, attention, and cache buffers for each drafter.

DFlash~\citep{chen2026dflash} captures its single parallel draft
forward. Domino~\citep{huang2026domino} captures its backbone,
output head, and complete GRU correction loop.
\method captures all $K$ passes and proposal construction
between passes. DSpark~\citep{cheng2026dspark} captures its
fixed-shape backbone, Markov correction, and confidence kernels;
its native confidence readback and dynamic width decision remain
in the measured host path. Target verification remains eager
for every method, as does the autoregressive baseline.

Draft and target GPU operations execute sequentially on the same
CUDA stream. During draft graph execution, however, the CPU can
enqueue the subsequent eager target operations. This enqueue-ahead
window can reduce gaps between target operations, partially hiding
the additional drafting overhead. Consequently, independently
measured draft and target latencies need not sum to the observed
round latency. Additional passes still incur GPU computation,
and their runtime benefit depends on the execution backend and
workload.

\subsubsection{Enqueue-Ahead Diagnostic}
\label{app:enqueue_ahead}

\begin{figure}[!htb]
  \centering
  \includegraphics[width=\linewidth]{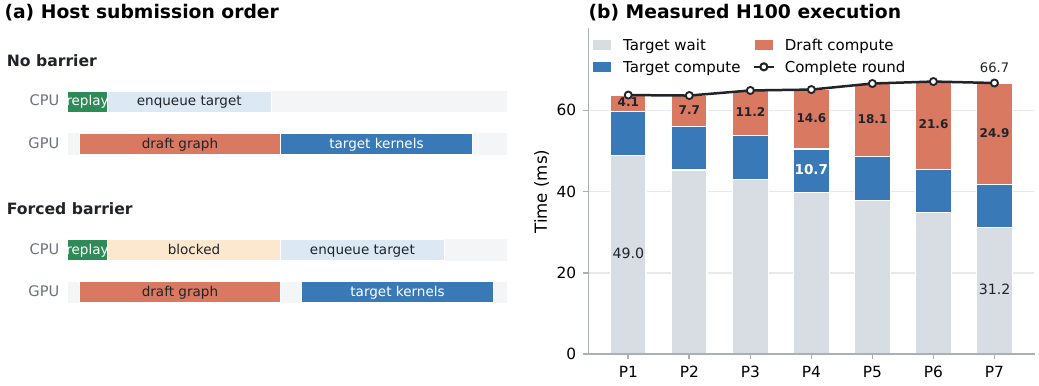}
  \caption{\textbf{Host enqueue-ahead during CUDA Graph replay.}
  (a) The CPU submits eager target operations while the GPU executes
  the captured drafter; draft and target GPU operations remain
  sequential. A forced host barrier delays target submission.
  (b) Nsight traces decompose each round into the pre-target draft
  interval, target compute, and gaps between target operations.}
  \label{fig:enqueue_ahead}
\end{figure}

In Figure~\ref{fig:enqueue_ahead}, all seven profiled settings execute the
same 2,294 target GPU
operations. Target compute remains nearly constant, while longer
draft execution is accompanied by shorter inter-operation gaps.
The forced-barrier control delays target dispatch without changing
the draft graph or arithmetic, supporting the role of host
enqueue-ahead. Because Nsight instrumentation perturbs latency,
these traces illustrate the mechanism rather than quantify its contribution
to the main-table speedups.

\FloatBarrier
\Needspace{8\baselineskip}
\section{Limitations and Future Work}
\label{sec:limitations}

\paragraph{Limitations.}
Our wall-clock gains are measured in a single-request setting with
CUDA Graph-optimized drafting and eager target verification.
CPU enqueue-ahead partially hides drafting overhead in this setup,
so these results do not directly establish speedups in production
serving with continuous batching. Additional editing passes still incur
computation, and their exposed cost can outweigh the benefit of
longer accepted prefixes. We evaluate targets of up to 8B parameters.
Because every drafter is trained from scratch for each target under a
matched budget, extending this controlled comparison to larger or newer
targets is left to future work.

\paragraph{Future Work.}
Realizing these acceptance gains in production requires joint
algorithm and scheduling design. A direct opportunity is to skip
editing passes that do not help: about two-thirds of editing passes
leave the accepted prefix unchanged
(Figure~\ref{fig:editing_mechanisms}(a)), yet each costs a drafter
forward. Predicting such rounds, for example
from proposal confidence or from whether predictions stop changing
across passes, could reduce drafting cost with little loss in acceptance.
The same signals could inform how to adapt proposal width and
verification length, and drafting and verification could be interleaved
across requests. Evaluating whether
these signals support effective compute allocation under realistic
serving workloads remains future work.

\end{document}